\documentclass[runningheads]{llncs}

\usepackage{eccv}

\usepackage{eccvabbrv}

\usepackage{graphicx}
\usepackage{booktabs}

\usepackage[accsupp]{axessibility}  

\usepackage{algpseudocode}
\usepackage{bbm}
\usepackage[linesnumbered,ruled,vlined]{algorithm2e}
\usepackage{caption}
\usepackage{subcaption}

\definecolor{dark_green}{rgb}{0, 0.5, 0}

\usepackage{multirow}

\newcommand{\hlrow}{\rowcolor{black!6}}

\usepackage{hyperref}

\usepackage{orcidlink}

\usepackage{amssymb}
\usepackage{pifont}
\usepackage{multirow}
\usepackage{diagbox}
\usepackage{colortbl}
\usepackage{lipsum}
\usepackage{wrapfig}
\usepackage{subcaption}
\newcommand{\ours}{LocoTrack\xspace}

\begin{document}

\title{Local All-Pair Correspondence for Point Tracking}


\author{Seokju Cho\inst{1} \and
Jiahui Huang\inst{2} \and
Jisu Nam\inst{1} \and
Honggyu An\inst{1} \and\\
Seungryong Kim\inst{1}\textsuperscript{,$\dagger$} \and
Joon-Young Lee\inst{2}\textsuperscript{,$\dagger$}
}

\authorrunning{S. Cho et al.}


\institute{\textsuperscript{1} Korea University\quad \textsuperscript{2} Adobe Research}

\maketitle
\begin{figure}
  \centering
   \resizebox{0.84\textwidth}{!}{\begin{minipage}{\textwidth}
  \includegraphics[width=\linewidth]{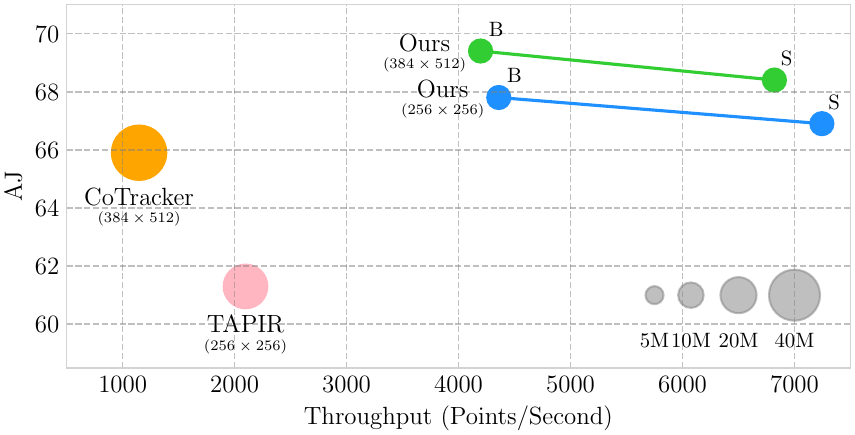}
  \end{minipage}}
\captionsetup{width=.84\linewidth}
\vspace{-5pt}
  \caption{\textbf{Evaluating \ours against state-of-the-art methods.} We compare our \ours against other SOTA methods~\cite{karaev2023cotracker,doersch2023tapir} in terms of model size (circle size), accuracy (y-axis), and throughput (x-axis). \ours shows exceptionally high precision and efficiency.}

  \label{fig:teaser}
  
  \vspace{-10pt}
\end{figure}

\begin{abstract}
We introduce \ours, a highly accurate and efficient model designed for the task of tracking any point (TAP) across video sequences. Previous approaches in this task often rely on local 2D correlation maps to establish correspondences from a point in the query image to a local region in the target image, which often struggle with homogeneous regions or repetitive features, leading to matching ambiguities. \ours overcomes this challenge with a novel approach that utilizes all-pair correspondences across regions, \ie, local 4D correlation, to establish precise correspondences, with bidirectional correspondence and matching smoothness significantly enhancing robustness against ambiguities. We also incorporate a lightweight correlation encoder to enhance computational efficiency, and a compact Transformer architecture to integrate long-term temporal information. \ours achieves unmatched accuracy on all TAP-Vid benchmarks and operates at a speed almost 6\(\times\) faster than the current state-of-the-art.
\end{abstract}

\let\thefootnote\relax\footnotetext{\textsuperscript{$\dagger$}Co-corresponding authors.}

\section{Introduction}
\begin{figure}[t]
  \centering
  \includegraphics[width=1.0\linewidth]{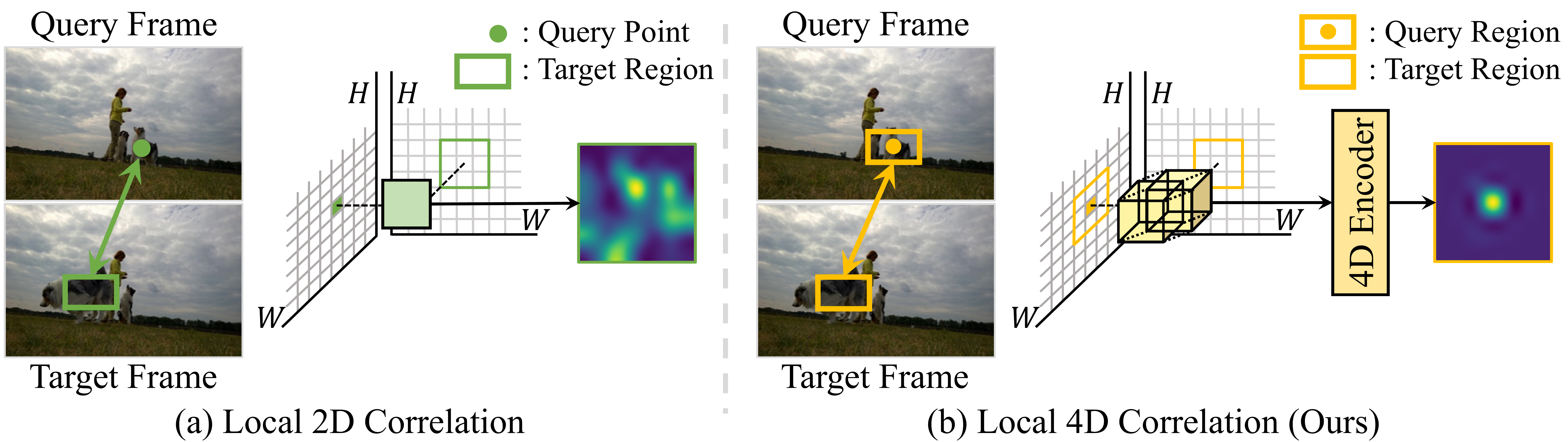}
  \vspace{-10pt}
  \caption{\textbf{Illustration of our core component.} Our local all-pair formulation, achieved with local 4D correlation, demonstrates robustness against matching ambiguity. This contrasts with previous works~\cite{harley2022particle,doersch2023tapir,karaev2023cotracker,vecerik2023robotap} that rely on point-to-region correspondences, achieved with local 2D correlation, which are susceptible to the ambiguity.}
  \label{fig:core-intuition}
\vspace{-10pt}
\end{figure}
Finding corresponding points across different views of a scene, a process known as point correspondence~\cite{lowe2004distinctive,bay2006surf,yi2018learning}, is one of fundamental problems in computer vision, which has a variety of applications such as 3D reconstruction~\cite{schonberger2016structure,mildenhall2021nerf}, autonomous driving~\cite{pollefeys2008detailed,janai2020computer}, and pose estimation~\cite{schonberger2016structure, sarlin2019coarse,sarlin2020superglue}. Recently, the emerging point tracking task~\cite{doersch2022tap, harley2022particle} addresses the point correspondence across a video. Given an input video and a query point on a physical surface, the task aims to find the corresponding position of the query point for every target frame along with its visibility status. This task demands a sophisticated understanding of motion over time and a robust capability for matching points accurately.

Recent methods in this task often rely on constructing a 2D local correlation map~\cite{harley2022particle, doersch2023tapir, vecerik2023robotap, karaev2023cotracker}, comparing the deep features of a query point with a local region of the target frame to predict the corresponding positions. 
However, this approach encounters substantial difficulties in precisely identifying positions within homogeneous areas, regions with repetitive patterns, or differentiating among co-occurring objects~\cite{yi2018learning,truong2020gocor,rocco2018neighbourhood}. To resolve matching ambiguities that arise in these challenging scenarios, establishing effective correspondence between frames is crucial. Existing works attempt to resolve these ambiguities by considering the temporal context~\cite{harley2022particle,doersch2023tapir,karaev2023cotracker,zheng2023pointodyssey}, however, in cases of severe occlusion or complex scenes, challenges often persist.

In this work, we aim to alleviate the problem with better spatial context which is lacking in local 2D correlations. We revisit dense correspondence methods~\cite{liu2010sift,rocco2017convolutional,truong2020glu, cho2021cats, cho2022cats++, truong2023pdc}, as they demonstrate robustness against matching ambiguity by leveraging rich spatial context. Dense correspondence establishes a corresponding point for every point in an image. To achieve this, these methods often calculate similarities for every pair of points across two images, resulting in a 4D correlation volume~\cite{rocco2017convolutional,truong2020glu,truong2020gocor,cho2021cats,nam2023diffmatch}.
This high-dimensional tensor provides dense bidirectional correspondence, offering matching priors that 2D correlation does not, such as dense matching smoothness from one image to another and vice versa. For example, 4D correlation can provide the constraint that the correspondence of one point to another image is spatially coherent with the correspondences of its neighboring points~\cite{rocco2018neighbourhood}. However, incorporating the advantages of dense correspondence, which stem from the use of 4D correlation, into point tracking poses significant challenges. Not only does it introduce a substantial computational burden but the high-dimensionality of the correlation also necessitates a dedicated design for proper processing~\cite{rocco2018neighbourhood,min2021hypercorrelation,cho2021cats}.

We solve the problem by formulating point tracking as a local all-pair correspondence problem, contrary to predominant point-to-region correspondence methods~\cite{harley2022particle,karaev2023cotracker,doersch2023tapir,vecerik2023robotap}, as illustrated in Fig.~\ref{fig:core-intuition}. We construct a local 4D correlation that finds all-pair matches between the local region around a query point and a corresponding local region on the target frame. With this formulation, our framework gains the ability to resolve matching ambiguities, provided by 4D correlation, while maintaining efficiency due to a constrained search range. The local 4D correlation is then processed by a lightweight correlation encoder carefully designed to handle high-dimensional correlation volume. This encoder decomposes the processing into two branches of 2D convolution layers and produces a compact correlation embedding. We then use a Transformer~\cite{karaev2023cotracker} to integrate temporal context into the embeddings. The Transformer's global receptive field facilitates effective modeling of long-range dependencies despite its compact architecture. Our experiments demonstrate that stack of 3 Transformer layers is sufficient to significantly outperform state-of-the-arts~\cite{doersch2023tapir,karaev2023cotracker}. Additionally, we found that using relative position bias~\cite{press2021train, raffel2020exploring, shaw2018self} allows the Transformer to process sequences of variable length. This enables our model to handle long videos without the need for a hand-designed chaining process~\cite{harley2022particle, karaev2023cotracker}.

Our model, dubbed \ours, outperforms the recent state-of-the-art model while maintaining an extremely lightweight architecture, as illustrated in Fig.~\ref{fig:teaser}. Specifically, our small model variant achieves a +2.5 AJ increase in the TAP-Vid-DAVIS dataset compared to Cotracker~\cite{karaev2023cotracker} and offers 6\(\times\) faster inference speed. Additionally, it surpasses TAPIR~\cite{doersch2023tapir} by +5.6 AJ with 3.5\(\times\) faster inference in the same dataset. 
Our larger variant, while still faster than competing state-of-the-art models~\cite{doersch2023tapir,karaev2023cotracker}, demonstrates even further performance gains.

In summary, \ours is a highly efficient and accurate model for point tracking. Its core components include a novel local all-pair correspondence formulation, leveraging dense correspondence to improve robustness against matching ambiguity, a lightweight correlation encoder that ensures computational efficiency, and a Transformer for incorporating temporal information over variable context lengths. 

\section{Related Work}
\subsubsection{Point correspondence.} The aim of point correspondence, which is also known as sparse feature matching~\cite{lowe2004distinctive,dusmanu2019d2, detone2018superpoint,sarlin2020superglue}, is to identify corresponding points across images within a set of detected points. This is often achieved by matching a hand-designed descriptors~\cite{bay2006surf, lowe2004distinctive} or, more recently, learnable deep features~\cite{detone2018superpoint, manuelli2020keypoints, sun2021loftr, jiang2021cotr}. They are also applicable to videos~\cite{pollefeys2008detailed}, as the task primarily targets image pairs with large baselines, which is similar to the case with video frames. These approaches filter out noisy correspondences using geometric constraints~\cite{hartley2003multiple, schonberger2016structure, torr1999feature} or their learnable counterparts~\cite{yi2018learning, sarlin2020superglue, jiang2021cotr}. However, they often struggle with objects that exhibit deformation~\cite{xiao2004closed}. Also, they primarily target the correspondence of geometrically salient points (\ie, detected points) rather than any arbitrary point.

\vspace{-10pt}
\subsubsection{Long-range point correspondence in video.} Recent methods~\cite{harley2022particle,doersch2022tap,doersch2023tapir,vecerik2023robotap,karaev2023cotracker,zheng2023pointodyssey,bian2023context} finds point correspondence in a video, aiming to find a track for a query point over a long sequence of video. They capture a long-range temporal context with MLP-Mixer~\cite{harley2022particle,bian2023context}, 1D convolution~\cite{doersch2023tapir, zheng2023pointodyssey}, or Transformer~\cite{karaev2023cotracker}. However, they either leverage a constrained length of sequence within a local temporal window and use sliding window inference to process videos longer than the fixed window size~\cite{harley2022particle,karaev2023cotracker,bian2023context}, or they necessitate a series of convolution layers to expand the temporal receptive field~\cite{doersch2023tapir,zheng2023pointodyssey}. Recent Cotracker~\cite{karaev2023cotracker} leverage spatial context by aggregating supporting tracks with self-attention. However, this approach requires tracking additional query points, which introduces significant computational overhead. Notably, Context-PIPs~\cite{bian2023context} constructs a correlation map across sparse points around the query and the target region. However, this sparsity may limit the model's ability to fully leverage the matching prior that all-pair correlation can provide, such as matching smoothness.

\vspace{-10pt}
\subsubsection{Dense correspondence.} Dense correspondence~\cite{liu2010sift} aims to establish pixel-wise correspondence between a pair of images. Conventional methods~\cite{melekhov2019dgc, truong2020glu, rocco2020efficient, cho2021cats, truong2023pdc, truong2021learning, nam2023diffmatch, teed2020raft, hong2024unifying} often leverage a 4-dimensional correlation volume, which computes pairwise cosine similarity between localized deep feature descriptors from two images, as the 4D correlation provides a mean for disambiguate the matching process. Traditionally, bidirectional matches from 4D correlation are filtered to remove spurious matches using techniques such as the second nearest neighbor ratio test~\cite{lowe2004distinctive} or the mutual nearest neighbor constraint. Recent methods instead learn patterns within the correlation map to disambiguate matches.
DGC-Net~\cite{melekhov2019dgc} and GLU-Net~\cite{truong2020glu} proposed a coarse-to-fine architecture leveraging global 4D correlation followed by local 2D correlation. CATs~\cite{cho2021cats,cho2022cats++} propose a transformer-based architecture to aggregate the global 4D correlation. GoCor~\cite{truong2020gocor}, NCNet~\cite{rocco2020efficient}, and RAFT~\cite{teed2020raft} developed an efficient framework using local 4D correlation to learn spatial priors in both image pairs, addressing matching ambiguities. 

The use of 4D correlation extends beyond dense correspondence. It has been widely applied in fields such as video object segmentation~\cite{oh2019video,cheng2022xmem}, few-shot semantic segmentation~\cite{min2021hypercorrelation,hong2022cost}, and few-shot classification~\cite{kang2021relational}. However, its application in point tracking remains underexplored. Instead, several attempts have been made to integrate the strengths of off-the-shelf dense correspondence model~\cite{teed2020raft} into point tracking. These include chaining dense correspondences~\cite{teed2020raft,harley2022particle}, which has limitations in recovering from occlusion, or directly finding correspondences with distant frames~\cite{neoral2024mft, wang2023tracking, moing2023dense}, which is computationally expensive.

\begin{figure}[t]
  \centering
  \includegraphics[width=1.0\linewidth]{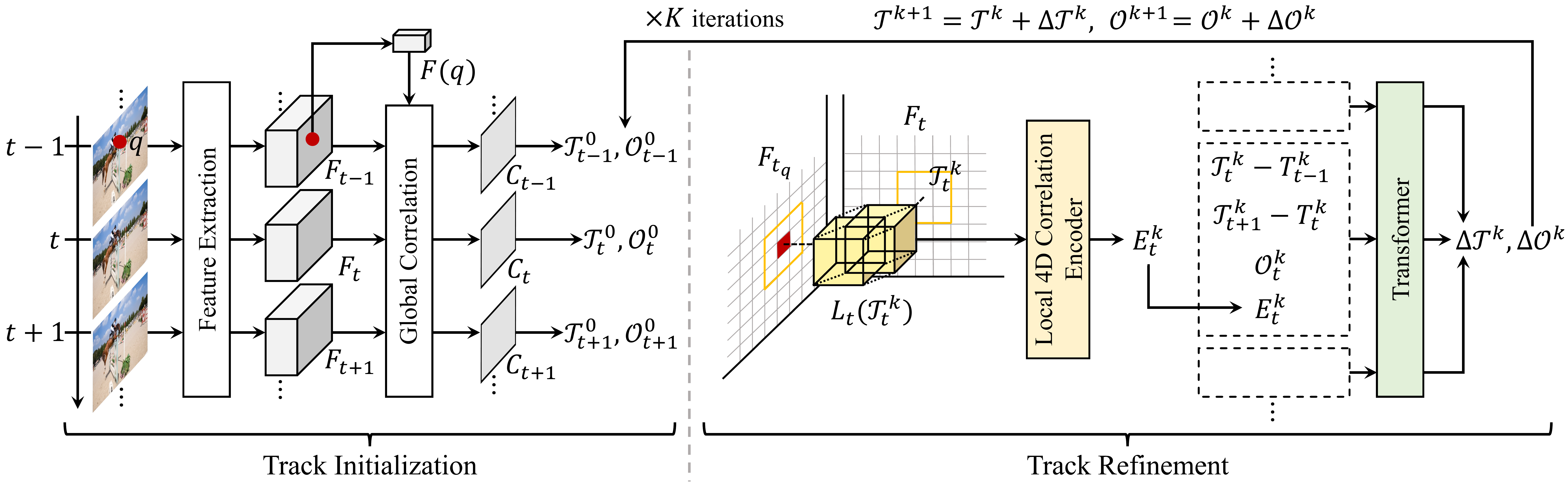}
  
  \vspace{-5pt}
  \caption{\textbf{Overall architecture of \ours.} Our model comprises two stages: track initialization and track refinement. The track initialization stage determines a rough position by conducting feature matching with global correlation. The track refinement stage iteratively refines the track by processing the local 4D correlation.}
  \label{fig:overall}
\vspace{-10pt}
\end{figure}

\section{Method}
In this work, we integrate the effectiveness of a 4D correlation volume into our point tracking pipeline. Compared to the widely used 2D correlation~\cite{harley2022particle,doersch2022tap,doersch2023tapir,karaev2023cotracker}, 4D correlation offers two distinct characteristics that provide valuable information for filtering out noisy correspondences, leading to more robust tracking:
\begin{itemize}
    \item \textbf{Bidirectional correspondence}: 4D correlation provides bidirectional correspondences, which can be used to verify matches and reduce ambiguity~\cite{lowe2004distinctive}. 
    This prior is often leveraged by checking for mutual consensus~\cite{rocco2018neighbourhood} or by employing a ratio test~\cite{lowe2004distinctive}.
    \item \textbf{Smooth matching}: A 4D correlation volume is constructed using dense all-pair correlations, which can be leveraged to enforce matching smoothness and improve matching consistency across neighboring points~\cite{rocco2018neighbourhood,truong2020glu,truong2020gocor}.
\end{itemize}
We aim to leverage these benefits of the 4D correlation volume while maintaining efficient computation. We achieve this by restricting the search space to a local neighborhood when constructing the 4D correlation volume. Along with the use of local 4D correlation, we also propose a recipe to benefit from the global receptive field of Transformers for long-range temporal modeling. This enables our model to capture long-range context within a few (even only with 3) stacks of transformer layers, resulting in a compact architecture. 

Our method, dubbed \ours, takes as input a query point \( q = (x_q, y_q, t_q) \in \mathbb{R}^3 \) and a video \( \mathcal{V} = \{ \mathcal{I}_t \}_{t=0}^{t=T-1} \), where \( T \) indicates the number of frames and \( \mathcal{I}_t \in \mathbb{R}^{H \times W \times 3} \) represents the \( t \)-th frame. We assume query point can be given in the arbitrary time step. Our goal is to produce a track \( \mathcal{T} = \{ \mathcal{T}_t \}_{t=0}^{t=T-1} \), where \( \mathcal{T}_t \in \mathbb{R}^2 \), and associated occlusion probabilities \( \mathcal{O} = \{ \mathcal{O}_t \}_{t=0}^{t=T-1} \), where \( \mathcal{O}_t \in [0, 1] \).
Following previous works~\cite{doersch2023tapir,karaev2023cotracker}, our method predicts the track in two stage approach: an initialization stage followed by a refinement stage, each detailed in the follows, as illustrated in Fig.~\ref{fig:overall}.

\begin{figure}[t]
  \centering
  \includegraphics[width=1.0\linewidth]{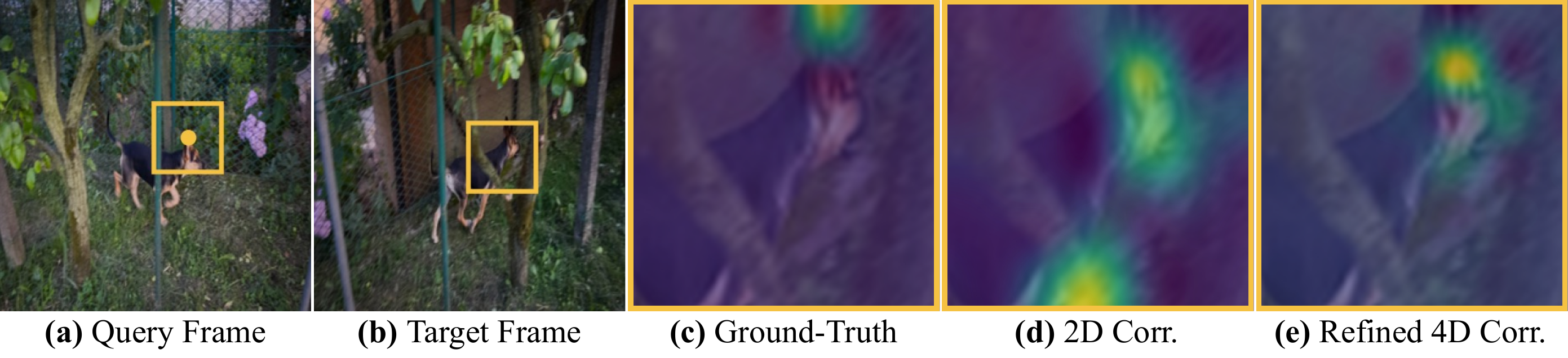}
  \vspace{-15pt}
  \caption{\textbf{Visualization of correspondence.} We visualize the correspondences established between the query and target regions. Our refined 4D correlation \textbf{(e)} demonstrates a clear reduction in matching ambiguity and yields better correspondences compared to the noisy results produced by 2D correlation \textbf{(d)}. This improvement aligns closely with the ground truth \textbf{(c)}. 
  }
  \label{fig:attention}
\vspace{-10pt}
\end{figure}
\subsection{Stage I: Track Initialization}\label{sec:track-initilization}
To estimate the initial track of a given query point, we conduct feature matching that constructs a global similarity map between features derived from the query point and the target frame's feature map, and choose the positions with the highest scores as the initial track. This similarity map, often referred to as a correlation map, provides a strong signal for accurately initializing the track's positions. We use a global correlation map for the initialization stage, which calculates the similarity for every pixel in each frame.

Specifically, we use hierarchical feature maps derived from the feature backbone~\cite{he2016deep}. Given a set of pyramidal feature maps \(\{F_t^l\}^{T-1}_{t} = \mathcal{E}(\mathcal{V})\), where \(\mathcal{E}(\cdot)\) represents the feature extractor and \(F^l_t\) indicates a level \(l\in \{0, \ldots, L - 1\}\) feature map in frame \(t\), we sample a query feature vector \(F^l(q)\) at position \(q\) from \(F^l\) using linear interpolation for each level \(l\). The global correlation map is calculated as \( \mathrm{C}^{l}_{t}=\frac{F_t^l\cdot F^l(q)}{\lVert F_t^l \rVert_2 \lVert F^l(q) \rVert_2 } \in \mathbb{R}^{H^l \times W^l} \), where \( H^l\) and \( W^l\) denote the height and width of the feature map at the \(l\)-th level, respectively. 
The correlation maps obtained from multiple levels are resized to the largest feature map size and concatenated as \(\mathrm{C}_{t}  \in \mathbb{R}^{H^0 \times W^0 \times L}\). The concatenated maps are processed as follows to generate the  initial track and occlusion probabilities:
\begin{align}
  \mathcal{T}_t^0 &= \mathrm{Softargmax}\left(\mathrm{Conv2D}\left( \mathrm{C}_t \right); \tau \right), \nonumber\\
  \mathcal{O}_t^0 &= \mathrm{Linear}([\mathrm{Maxpool}(\mathrm{C}_t);\ \mathrm{Avgpool}(\mathrm{C}_t)]),
\end{align}
where \(\mathrm{Conv2D}: \mathbb{R}^{H\times W \times L} \rightarrow \mathbb{R}^{ H\times W}\) is a single-layered 2D convolution layer, \(\mathrm{Softargmax}: \mathbb{R}^{H\times W} \rightarrow \mathbb{R}^2\) is a differentiable argmax function with a Gaussian kernel~\cite{lee2019sfnet} that provides the 2D position of the maximum value, \(\tau\) is a temperature parameter, \([\cdot]\) indicates concatenation, and \(\mathrm{Linear}: \mathbb{R}^{2L} \rightarrow \mathbb{R}\) is a linear projection. Similar to CBAM~\cite{woo2018cbam}, we apply global max and average pooling followed by a linear projection to calculate initial occlusion probabilities. 

\subsection{Stage II: Track Refinement}\label{sec:track-refinement}
We found that the initial track \(\mathcal{T}^0\) and \(\mathcal{O}^0\) often exhibit severe jittering, arising from the matching ambiguity from the noisy correlation map. 
We iteratively refine the noise in the initial tracks \(\mathcal{T}^0\) and \(\mathcal{O}^0\). For each iteration, we estimate the residuals \(\Delta\mathcal{T}^k\) and \(\Delta\mathcal{O}^k\), which are then applied to the tracks as \(\mathcal{T}^{k+1} := \mathcal{T}^k + \Delta\mathcal{T}^k\) and \(\mathcal{O}^{k+1} := \mathcal{O}^k + \Delta\mathcal{O}^k\). During the refining process, the matching noise can be rectified in two ways: 1) by establishing locally dense correspondences with local 4D correlation, and 2) through temporal modeling with a Transformer~\cite{vaswani2017attention}.

\vspace{-10pt}
\subsubsection{Local 4D correlation.} 
The 2D correlation \(\mathcal{C}_t\) often exhibits limitations when dealing with repetitive patterns or homogeneous regions as exemplified in Fig.~\ref{fig:attention}. Inspired by dense correspondence literatures, we utilize 4D correlation to provide richer information for refining tracks compared to 2D correlation. The 4D correlation \(\mathrm{C}^{\mathrm{4D}} \in \mathbb{R}^{H\times W \times H\times W}\), which computes every pairwise similarity, can be formally defined as follows:
\begin{gather}
\mathrm{C}^{\mathrm{4D}}_{t}(i,j)=\frac{F_t(i)\cdot F_{t_q}(j)}{\lVert F_t(i) \rVert_2 \lVert F_{t_q}(j) \rVert_2 },
\end{gather}
where \(F_{t_q}\) is the feature map from the frame in which the query point is located, and \(i\) and \(j\) specify the locations within the feature map. However, since a global 4D correlation volume with the shape of \(H \times W \times H \times W\) becomes computationally intractable, we employ a local 4D correlation \(\mathrm{L} \in \mathbb{R}^{h_p \times w_p \times h_q \times w_q}\), where \((h_p, w_p, h_q, w_q)\) denotes spatial resolution of local correlation. We define the correlation as follows:
\begin{gather}
\mathcal{N}(p, r) = \{ p + \delta \mid \delta\in \mathbb{Z}^2, \lVert\delta \rVert_{\infty} \leq r \}, \nonumber\\
\mathrm{L}_{t}(i,j;p) =  \frac{F_t(i)\cdot F_{t_q}(j)}{\lVert F_t(i) \rVert_2 \lVert F_{t_q}(j) \rVert_2 }  ,\quad i \in \mathcal{N}(p;r_p), \quad j \in \mathcal{N}(q;r_q),
\end{gather}
where \(r_p\) and \(r_q\) are the radii of the regions around points \(p\) and \(q\), respectively, resulting in \(h_p=w_p=2r_p+1\) and \(h_q=w_q=2r_q + 1\). 
The correlation then serves as a cue for refining the track  \(\mathcal{T}^k\). To achieve this, we calculate the set of local correlations around the intermediate predicted position, denoted as \(\{ \mathrm{L}_{t}({\mathcal{T}^k_t}) \}_{t=0}^{T-1}\) with abuse of notation.
\vspace{-10pt}

\begin{wrapfigure}{r}{0.35\textwidth}
\centering
    \includegraphics[width=0.8\linewidth]{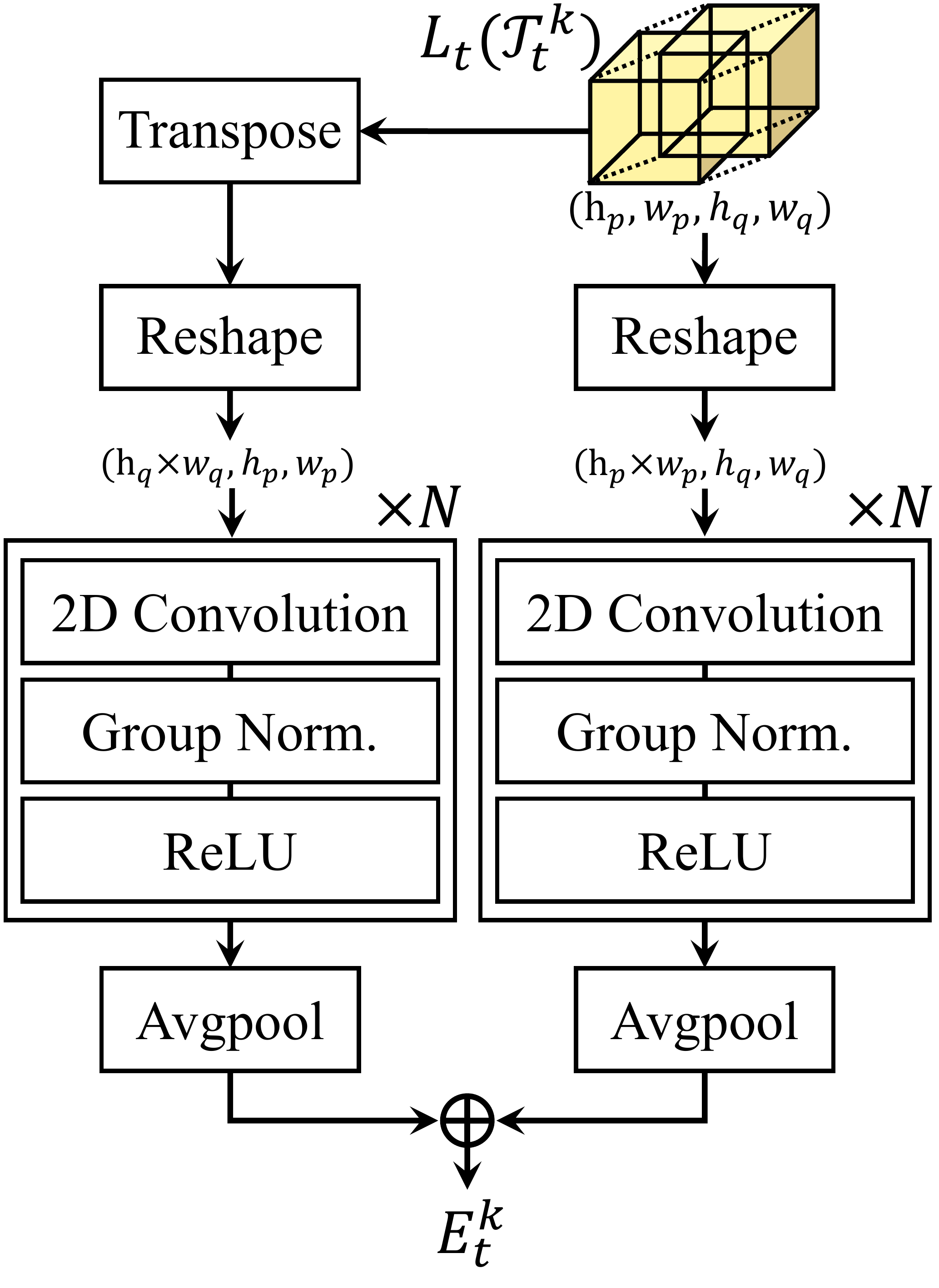}
    \\
    \vspace{-5pt}
    \captionsetup{width=.9\linewidth}
    \caption{\textbf{Local 4D correlation encoder.} }
    \label{fig:corr_encoder}
\end{wrapfigure}

\subsubsection{Local 4D correlation encoder.} We then process the local 4D correlation volume to disambiguate matching ambiguities, leveraging the smoothness of both the query and target dimensions of correlations. Note that the obtained 4D correlation is a high-dimensional tensor, posing an additional challenge for its correct processing. In this regard, we introduce an efficient encoding strategy that decomposes the processing of the correlation.  We process the 4D correlation in two symmetrical branches as shown in Fig.~\ref{fig:corr_encoder}. One branch spatially processes the dimensions of the query, treating the flattened target dimensions as a channel dimension. The other branch, on the other hand, considers the query dimensions as channel. Each branch compresses the correlation into a single vector, which are then concatenated to form a correlation embedding \(E_t^k\):
\begin{align}
E_t^k = \left[\mathcal{E}_\mathrm{L} \left(\mathrm{L}_t\left(\mathcal{T}^k_t\right) \right);\ \mathcal{E}_\mathrm{L} \left( \left(\mathrm{L}_t \left(\mathcal{T}^k_t\right) \right)^T \right) \right],
\end{align}
where \( \mathrm{L}(i, j) = \mathrm{L}^T(j, i) \).
The convolutional encoder \(\mathcal{E}_\mathrm{L}: \mathbb{R}^{h_p \times w_p \times h_q \times w_q} \rightarrow \mathbb{R}^{C_E} \) consists of stacks of strided 2D convolutions, group normalization~\cite{wu2018group}, and ReLU activations. These operations progressively reduce the correlation's spatial dimensions, followed by a final average pooling layer for a compact representation. We obtain the correlation embedding for all feature levels \(l\), and concatenate them to form the final embedding. For more details on the local 4D correlation encoder, please refer to the supplementary material.
\vspace{-10pt}

\subsubsection{Temporal modeling with length-generalizable transformer.} The encoded correlation is then provided to the refinement model. The model refines the initial trajectory and predicts its error with respect to the ground truth, \(\Delta\mathcal{T}\) and \(\Delta\mathcal{O}\), which requires an ability to leverage temporal context. For the temporal modelling, we explore three candidates widely used in the literature: 1D Convolution~\cite{doersch2023tapir,zheng2023pointodyssey}, MLP-Mixer~\cite{harley2022particle}, and Transformer~\cite{karaev2023cotracker}. We consider two aspects to select the appropriate architecture: 1) Can the architecture handle arbitrary sequence lengths \(T\) at test time? 2) Can the temporal receptive field, crucial for capturing long-range context, be sufficiently large with just a few layers stacked? Based on these criteria, we choose the Transformer as our architecture because it can handle arbitrary sequence lengths, a capability the MLP-Mixer lacks. This lack would necessitate an additional test-time strategy (\eg, sliding window inference~\cite{harley2022particle}) to accommodate sequences longer than those used during training. Additionally, the Transformer can form a global receptive field with a single layer, unlike convolution, which requires multiple layers to achieve an expanded receptive field.

Although the Transformer can process sequences of arbitrary length at test time, we found that sinusoidal position encoding~\cite{vaswani2017attention} degrades performance for videos with sequence lengths that differ from those used during training. Instead, we use relative position bias~\cite{press2021train, raffel2020exploring, shaw2018self}, which disproportionately reduces the impact of distant tokens by adjusting the bias within the Transformer's attention map.
However, relative position bias is based solely on the distance between tokens cannot distinguish their relative direction (\eg, whether token A is before or after token B), which makes it only suitable for causal attention. To address this, we divide the attention head into two groups: one group encodes relative position only for tokens on the left, and the other for tokens on the right:
\begin{align}
&\mathrm{Softmax}(\mathrm{q}\cdot \mathrm{k}^T + b(h)), \text{ where} \nonumber\\
&b(t_1, t_2;h) = \begin{cases}
b_{\mathrm{left}}(t_1, t_2; h), & h < \left\lfloor \frac{N_h}{2} \right\rfloor, \\
b_{\mathrm{right}}(t_1, t_2; h - \lfloor N_h / 2\rfloor), & h \ge \left\lfloor \frac{N_h}{2} \right\rfloor,
\end{cases}
\end{align}
where \(\mathrm{q}\) and \(\mathrm{k}\) denote the query and key, respectively, \(N_{\mathrm{h}}\) is the number of heads, and \(h \in \{0,...,N_{\mathrm{h}}-1 \}\) is the index of the attention head. The bias term \(b_{\mathrm{left}}\) adjusts the attention map to ensure that each query token attends only to key tokens located to its left or within the same position, as follows:
\begin{align}
b_{\mathrm{left}}(t_1, t_2;h)= \begin{cases}
-\infty, & \text{if } t_1 < t_2, \\
-s_h| t_1 - t_2 |, & \text{if } t_1 \geq t_2, \\
\end{cases}
\end{align}
where \(s_h\in \mathbb{R}^+\) is a scaling factor that controls the rate of bias decay as distance increases. We employ different scaling factors for each attention head, following Press et al.~\cite{press2021train}. The function \(b_{\mathrm{right}}(\cdot)\) can be similarly defined. With this design choice, we found that the Transformer can generalize to videos of arbitrary length, eliminating the need for test-time hand-designed techniques such as sliding window inference~\cite{harley2022particle,karaev2023cotracker}. 

\vspace{-10pt}
\subsubsection{Iterative update.} We stack \(N_S\) Transformer layers with the modified self-attention and feed the correlation embedding \(\{E_t^k\}_{t=0}^{t=T-1}\), the encoded initialized track \(\mathcal{T}^k\), and occlusion status \(\mathcal{O}^k\) to the Transformer \(\mathcal{E}_S\) to predict track updates. We found using position differences between adjacent frames improves training convergence compared to using the absolute positions. This is formally defined as:
\begin{gather}
\Delta\mathcal{T}^k, \Delta\mathcal{O}^k = \mathcal{E}_S\left( \left\{ \left[ \sigma \left(\mathcal{T}^k_t - \mathcal{T}^k_{t - 1}\right);\ \sigma\left(\mathcal{T}^k_{t + 1} - \mathcal{T}^k_{t}\right) ;\ \mathcal{O}^k_t ;\ E_t^k\right] \right\}_{t=0}^{t=T-1}\right),\nonumber\\
\mathcal{T}^k_{-1} := \mathcal{T}^k_{0}, \qquad \mathcal{T}^k_{T} := \mathcal{T}^k_{T-1},
\end{gather}
where \(\sigma(\cdot)\) is a sinusoidal encoding~\cite{tancik2020fourier}, \([\cdot]\) denotes concatenation, and \(\Delta\mathcal{T}^k\) and \(\Delta\mathcal{O}^k\) are predicted updates. Sequentially, the predicted updates are applied to initial track as \(\mathcal{T}^{k+1}:= \mathcal{T}^k + \Delta\mathcal{T}^k\) and \(\mathcal{O}^{k+1}:= \mathcal{O}^k + \Delta\mathcal{O}^k\). We perform \(K\) iterations, yielding the final refined track \(\mathcal{T}^K\) and \(\mathcal{O}^K\).

\section{Experiments}
\subsection{Implementation Details} 
We use JAX~\cite{jax2018github} for implementation. For training, we utilize the Panning MOVi-E dataset~\cite{doersch2023tapir} generated with Kubric~\cite{greff2022kubric}. We employ the loss functions introduced in Doersch et al.~\cite{doersch2023tapir}, including the prediction of additional uncertainty estimation for both track initialization and a refinement model. We use the AdamW~\cite{loshchilov2017decoupled} optimizer and use \(1\cdot 10^{-3}\) for both learning rate and weight decay. We employ a cosine learning rate scheduler with a \(1000\)-step warmup stage~\cite{loshchilov2016sgdr}. Following Sun et al.~\cite{sun2022disentangling}, we apply gradient clipping with a value of \(1.0\). The initialization stage is first trained for 100K steps, followed by track refinement model training for an additional 300K steps. This process takes approximately 4 days on 8 NVIDIA RTX 3090 GPUs with a batch size of \(1\) per GPU. For each batch, we randomly sample \(256\) tracks. We use a \(256\times256\) training resolution, following the standard protocol of TAP-Vid benchmark.

Our feature backbone is ResNet18~\cite{he2016deep} with instance normalization~\cite{ulyanov2016instance} replacing batch normalization~\cite{ioffe2015batch}. We use three pyramidal feature maps (\(L=3\)) from ResNet, each with a stride of 2, 4 and 8, respectively. The temperature value for softargmax is set to \(\sigma=20.0\). The radii of the local correlation window are \(r_q=r_p=3\). We stack \(N_S=3\) Transformer layers for \(\mathcal{E}_S\). The number of iterations (\(K\)) is set to 4. For the track refinement model, we propose two variants: a small model and a base model. All ablations are conducted using the base model. The hidden dimension of the Transformer is set to 256 for the small model and 384 for the base model. The number of heads is set to 4 for the small model and 6 for the base model. For more details, please refer to supplementary materials.

\begin{table}[t]
  \caption{\textbf{Quantitative comparison on the TAP-Vid datasets with the strided query mode.} Throughput is measured on a single Nvidia RTX 3090 GPU. 
  }
  \vspace{-10pt}
  \label{tab:strided-quan}
  \centering
\resizebox{\linewidth}{!}{
  \begin{tabular}{l|ccc|ccc|ccc|c}
    \toprule
\multirow{2}{*}{Method} & \multicolumn{3}{c|}{Kinetics} & \multicolumn{3}{c|}{DAVIS} & \multicolumn{3}{c|}{RGB-Stacking} & Throughput  \\
&  AJ & $<\delta^{x}_{avg}$ & OA &  AJ & $<\delta^{x}_{avg}$ & OA &  AJ & $<\delta^{x}_{avg}$ & OA & (points/sec)  \\
\midrule
\midrule
\multicolumn{2}{l}{\textit{Input Resolution 256\(\times\)256}}\\
Kubric-VFS-Like~\cite{greff2022kubric}& 40.5 & 59.0 & 80.0 & 33.1 & 48.5 & 79.4 & 57.9 & 72.6 & 91.9 & -\\ %
TAP-Net~\cite{doersch2022tap}& 46.6 & 60.9 & 85.0 &  38.4 & 53.1 & 82.3 & 59.9 & 72.8 & 90.4 & \textbf{29,535.98} \\ %
RAFT~\cite{teed2020raft}& 34.5 & 52.5 & 79.7 &  30.0 & 46.3 & 79.6 & 44.0 & 58.6 & 90.4 & \underline{23,405.71} \\ %

TAPIR~\cite{doersch2023tapir}& 57.2 & 70.1 & 87.8 &  61.3 & 73.6 & 88.8 & 62.7 & 74.6 & 91.6 & 2,097.32\\ %
\hlrow \ours-S  & \textbf{59.6} & \underline{72.7} & \underline{88.1} & \underline{66.9} & \underline{78.8} & \underline{88.9} & \textbf{77.4} & \textbf{87.0} & \underline{92.9}& 7,244.47 \\
\hlrow \ours-B  & \underline{59.5} & \textbf{73.0} & \textbf{88.5}  & \textbf{67.8} & \textbf{79.6} & \textbf{89.9} & \underline{77.1} & \underline{86.9} & \textbf{93.2} & 4,358.96 \\

 \midrule
\multicolumn{2}{l}{\textit{Input Resolution 384\(\times\)512}}\\
PIPs~\cite{harley2022particle}& 35.3 & 54.8 & 77.4 & 42.0 & 59.4 & 82.1 & 37.3 & 51.0 & \textbf{91.6} & 46.43 \\ %
FlowTrack~\cite{cho2024flowtrack}& - & - & - & 66.0 & 79.8 & 87.2 & - & - & - & -\\ %
CoTracker~\cite{karaev2023cotracker}& - & - & - &  65.9 & 79.4 & \textbf{89.9} & - & - & - & 1,146.79\\ %
 \hlrow \ours-S & \underline{58.7} & \underline{72.2} & \underline{84.5} & \underline{68.4} & \underline{80.4} & 87.5 & \textbf{71.0} & \textbf{84.4} & 83.3 & \textbf{6,820.57}  \\
 \hlrow \ours-B & \textbf{59.1} & \textbf{72.5} & \textbf{85.7} & \textbf{69.4} & \textbf{81.3} & \underline{88.6} & \underline{70.8} & \underline{83.2} & \underline{84.1} & \underline{4,196.36}  \\
  \bottomrule
  \end{tabular}}
  \vspace{-10pt}
\end{table}

\begin{table}[t]
  \caption{\textbf{Quantitative comparison on the query first mode.}
  }
  \vspace{-10pt}
  \label{tab:query-first}
  \centering
  \begin{tabular}{l|ccc|ccc|ccc}
    \toprule
 \multirow{2}{*}{Method} & \multicolumn{3}{c|}{Kinetics-First} & \multicolumn{3}{c|}{DAVIS-First} & \multicolumn{3}{c}{RoboTAP-First} \\ %
 & AJ & $<\delta^{x}_{avg}$ & OA & AJ & $<\delta^{x}_{avg}$ & OA & AJ & $<\delta^{x}_{avg}$ & OA \\
\midrule\midrule
\multicolumn{5}{l}{\textit{Input Resolution 256\(\times\)256}}\\
TAP-Net~\cite{doersch2022tap} & 38.5 & 54.4 & 80.6 & 33.0 & 48.6 & 78.8 & 45.1 & 62.1 & 82.9 \\
TAPIR~\cite{doersch2023tapir} & 49.6 & 64.2 & \underline{85.0} & 56.2 & 70.0 & \underline{86.5} & 59.6 & 73.4 & \underline{87.0} \\
\hlrow \ours-S & \underline{52.8} & \underline{66.5} & 84.9 & \underline{62.0} & \underline{74.3} & 86.1 & \textbf{62.5} & \underline{76.0} & \underline{87.0}\\
\hlrow \ours-B & \textbf{52.9} & \textbf{66.8} & \textbf{85.3} & \textbf{63.0} & \textbf{75.3} & \textbf{87.2} & \underline{62.3} & \textbf{76.2}& \textbf{87.1} \\
      
\midrule
\multicolumn{5}{l}{\textit{Input Resolution 384\(\times\)512}}\\
CoTracker~\cite{karaev2023cotracker} & 48.7 & 64.3 & \textbf{86.5} & 60.6 & 75.4 & \textbf{89.3} & - & - & - \\
\hlrow \ours-S & \underline{51.9} & \underline{66.1} & 81.2 & \underline{63.2} & \underline{76.2} & 84.6 & - & -& -\\
\hlrow \ours-B & \textbf{52.3} & \textbf{66.4} & \underline{82.1} & \textbf{64.8} & \textbf{77.4} & \underline{86.2} & - & - & - \\
\bottomrule
  \end{tabular}
  
\vspace{-15pt}
\end{table}
\begin{figure}[t]
  \centering
\includegraphics[width=\linewidth]{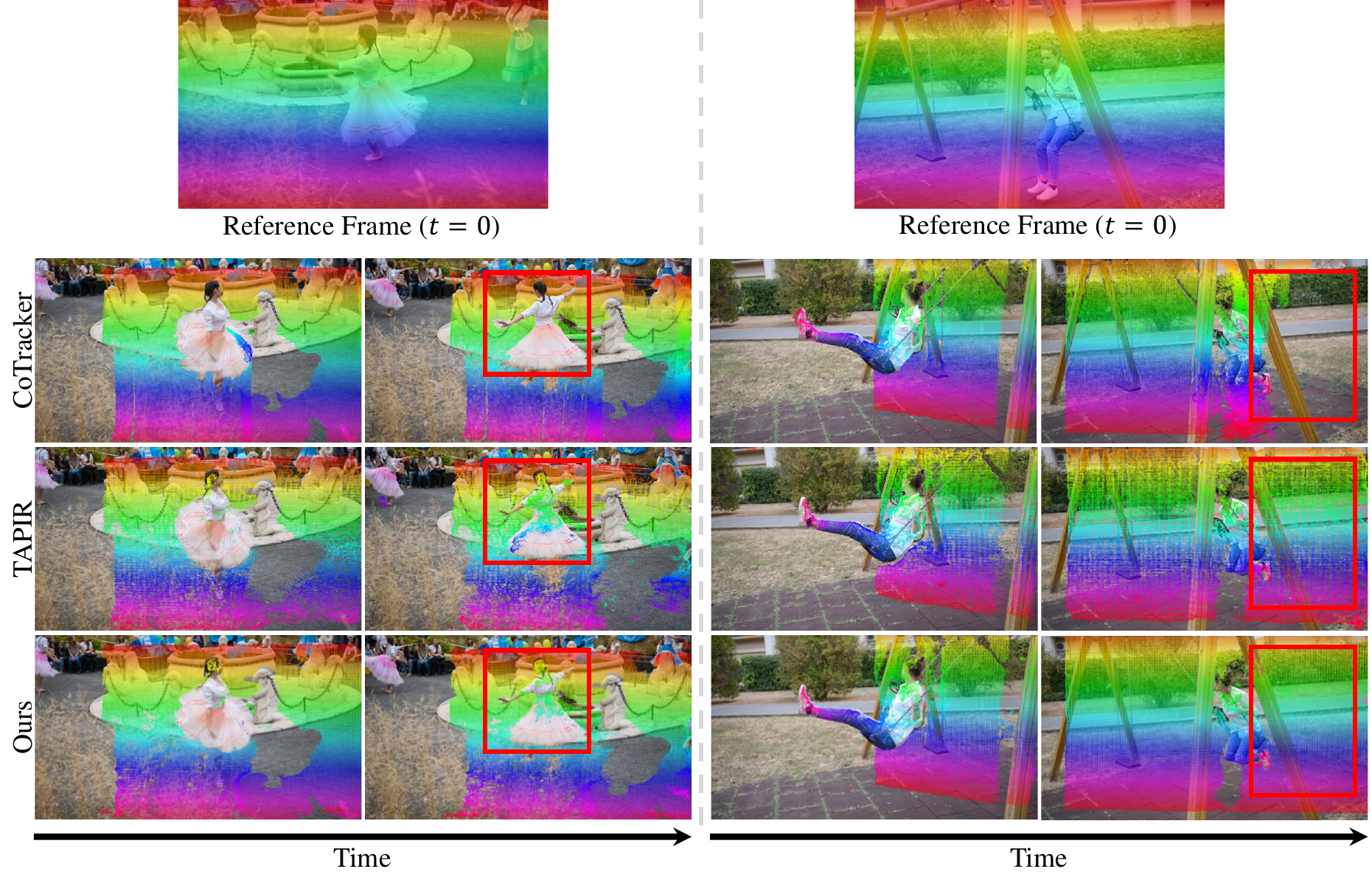}
\\
\vspace{-10pt}
\caption{\textbf{Qualitative comparison of long-range tracking.} We visualize dense tracking results generated by \ours and state-of-the-art methods~\cite{doersch2023tapir,karaev2023cotracker}. These visualizations use query points densely distributed within the initial reference frame. Our model can establish highly precise correspondences over long ranges, even in the presence of occlusions and matching challenges like homogeneous areas or deforming objects. Best viewed in color.}
\label{fig:qualitative}
\vspace{-10pt}
\end{figure}

\subsection{Evaluation Protocol} 
We evaluate the precision of the predicted tracks using the TAP-Vid benchmark~\cite{doersch2022tap} and the RoboTAP dataset~\cite{vecerik2023robotap}. 
For evaluation metrics, we use position accuracy (\(<\delta^x_{avg}\)), occlusion accuracy (OA), and average Jaccard (AJ). \(<\delta^x_{avg}\) calculates position accuracy for the points visible in ground-truth. It calculates the percentage of correct points (PCK)~\cite{rocco2018neighbourhood}, averaged over the error threshold values of 1, 2, 4, 8, and 16 pixels. OA represents the average accuracy of the binary classification results for occlusion. AJ is a metric that evaluates both position accuracy and occlusion accuracy.

Following Doersch et al.~\cite{doersch2022tap}, we evaluate the datasets in two modes: strided query mode and first query mode. Strided query mode samples the query point along the ground-truth track at fixed intervals, sampling every 5 frames, whereas first query mode samples the query point solely from the first visible point.

\begin{table}[t]
  \caption{\textbf{Comparison of computation cost.}  We measure the inference time with a varying number of query points and calculate the FLOPs for the feature backbone and refinement stage, along with the number of parameters. All metrics are measured using a video consisting of 24 frames on a single Nvidia RTX 3090 GPU. 
  }
  \vspace{-10pt}
  \label{tab:computation}
  \centering
  \resizebox{\linewidth}{!}{
  \begin{tabular}{l|cccccc|cccc}
    \toprule
\multirow{2}{*}{Method} & \multicolumn{6}{c|}{Inference Time (s)} & Throughput & Backbone & FLOPs & \# of \\ %
 & 10\(^0\) point & 10\(^1\) points & 10\(^2\) points & 10\(^3\) points & 10\(^4\) points & 10\(^5\) points  & (points/sec) & FLOPs (G) & per point (G) & Params. (M) \\
\midrule\midrule
RAFT~\cite{teed2020raft} & - & - & - & - & - & - & 23,405.71 & 325.45 & - & 5.3 \\
\midrule
CoTracker~\cite{karaev2023cotracker} & 0.53 & 0.53 & 0.53 & 1.18 & 8.40 & 87.2 & 1,146.79 & \underline{624.83} & 4.65 & 45.5 \\
TAPIR~\cite{doersch2023tapir} & \underline{0.06} & \underline{0.06} & 0.19 & 0.82 & 4.89 & 47.68 & 2,097.32 & \textbf{442.16} & 5.12 & 29.3 \\
\hlrow\ours-S  &  \textbf{0.04} & \textbf{0.04} & \textbf{0.05} & \textbf{0.17} & \textbf{1.44} & \textbf{14.23} & \textbf{7,244.47} & \textbf{442.16} & \textbf{1.08} & \textbf{8.2} \\
\hlrow\ours-B  &  \textbf{0.04} & \textbf{0.04} & \underline{0.06} & \underline{0.26} & \underline{2.39} & \underline{23.37} & \underline{4,358.96}  & \textbf{442.16} & \underline{2.10} & \underline{11.5} \\
\bottomrule
  \end{tabular}
  }
  \vspace{-10pt}
\end{table}
\subsection{Main Results}
\subsubsection{Quantitative comparison.} We compare our method with recent state-of-the-art approaches~\cite{greff2022kubric,doersch2022tap,doersch2023tapir,teed2020raft,harley2022particle,karaev2023cotracker,cho2024flowtrack} in both strided query mode, with scores shown in Table~\ref{tab:strided-quan}, and first query mode, with scores shown in Table~\ref{tab:query-first}. To ensure a fair comparison, we categorize models based on their input resolution sizes: \(256\times 256\) and  \(384\times 512\). 
Along with performance, we also present the throughput of each model, which indicates the number of points a model can process within a second. Higher throughput implies more efficient computation.

Our small variant, \ours-S, already achieves state-of-the-art performance on AJ and position accuracy across all benchmarks, surpassing both TAPIR and CoTracker by a large margin. In the DAVIS benchmark with strided query mode, we achieved a \(+\)5.6 AJ improvement compared to TAPIR and a \(+\)2.5 AJ improvement compared to CoTracker. This small variant model is not only powerful but also extremely efficient compared to recent state-of-the-art methods. Our model demonstrates 3.5\(\times\) higher throughput than TAPIR and 6\(\times\) higher than CoTracker. \ours-B model shows even better performance, achieving a \(+\)0.9 AJ improvement over our small variant in DAVIS strided query mode.

However, our model often shows degradation on some datasets in \(384\times 512\). We attribute this degradation to the diminished effective receptive field of local correlation when resolution is increased.

\vspace{-10pt}
\subsubsection{Qualitative comparison.} The qualitative comparison is shown in Fig.~\ref{fig:qualitative}. We visualize the results from the DAVIS~\cite{pont20172017} dataset, with the input resized to \(384\times 512\) resolution. Note that images at their original resolution are used for visualization. Overall, our method demonstrates superior smoothness compared to TAPIR. Our predictions are spatially coherent, even over long-range tracking sequences with occlusion.

\subsection{Analysis and Ablation Study}
\subsubsection{Efficiency comparison.} We compare efficiency to recent state of the arts~\cite{teed2020raft,karaev2023cotracker,doersch2023tapir} in Table~\ref{tab:computation}. We measure inference time, throughput, FLOPs, and the number of parameters for a 24 frame video. We report inference time for a varying number of query points, increasing exponentially from \(10^0\) to \(10^5\). To measure throughput, we calculate the average time required to add each query point. Also, we measure FLOPs for both the feature backbone and the refinement model, focusing on the incremental FLOPs per additional point. 

All variants of our model demonstrate superior efficiency across all metrics. Our small variant exhibits 4.7\(\times\) lower FLOPs per point compared to TAPIR and 4.3\(\times\) lower than CoTracker. Additionally, our model boasts a compact parameter count of only 8.2M, which is 5.5\(\times\) lower than CoTracker. Remarkably, our model can process \(10^4\) points in approximately one second, implying real-time processing of \(64\times 64\) near-dense query points for a 24 frame rate video. This underscores the practicality of our model, paving the way for real-time applications.  

\begin{table}[t]
  \setlength{\tabcolsep}{5pt}
  \caption{\textbf{Ablation on construction of correlation volume.}
  }
  \vspace{-10pt}
  \label{tab:local-correlation}
  \centering
  \resizebox{.7\textwidth}{!}{
  \begin{tabular}{lcc|ccc}
    \toprule
& Local Corr.& Query& \multicolumn{3}{c}{DAVIS} \\
& Size & Neighbour & AJ &$<\delta^{x}_{avg}$ & OA \\
\midrule\midrule
\textbf{(I)}& \(7 \times 7\) & No neighbour (2D corr.) & 65.0 & 77.2 & 89.0\\
\textbf{(II)}& \(9 \times 7 \times 7\) & Uniform random in local region & 65.7 & 77.8 & 88.9\\
\textbf{(III)}& \(1 \times 9 \times 7 \times 7\) & Horizontal line& 66.5 & 78.4 & 89.4\\
\textbf{(IV)}& \(3 \times 3 \times 7 \times 7\) & Regular grid (\(r_q=1\)) & \underline{67.2} & \underline{79.1} & \underline{89.5}\\
\hlrow \textbf{(V)}& \(7 \times 7 \times 7 \times 7\) & Regular grid (\(r_q=3\), Ours) & \textbf{67.8} & \textbf{79.6} & \textbf{89.9}\\
\bottomrule
  \end{tabular}
  }
  
\vspace{-10pt}
\end{table}

\begin{table}[t]
\parbox{.482\linewidth}{
  \caption{\textbf{Ablation on position encoding.} 
  }
  \vspace{-10pt}
  \label{tab:position-encoding}
  \centering
  \resizebox{\linewidth}{!}{
  \begin{tabular}{ll|ccc}
    \toprule
& \multirow{2}{*}{Method} & \multicolumn{3}{c}{DAVIS} \\
& & AJ &$<\delta^{x}_{avg}$ & OA\\
\midrule\midrule
\textbf{(I)}& Sinusoidal encoding~\cite{vaswani2017attention} & 61.9 & 73.9 & 83.5 \\
\hlrow\textbf{(II)}& Relative position bias (Ours) & \textbf{67.8} & \textbf{79.6} & \textbf{89.9}\\

\bottomrule
  \end{tabular}
  }
  }
\hfill
\parbox{0.504\linewidth}{
  \caption{\textbf{Ablation on architecture of \(\mathcal{E}_S\).} We found that our model outperforms its counterpart while using the same number of parameters.
  }
  \vspace{-10pt}
  \label{tab:architecture-ablation}
  \centering
  \resizebox{\linewidth}{!}{
  \begin{tabular}{ll|cc|ccc}
    \toprule
& \multirow{2}{*}{Method} & \# of & \# of & \multicolumn{3}{c}{DAVIS} \\
& & Layers & Params. & AJ &$<\delta^{x}_{avg}$ & OA\\
\midrule\midrule
\textbf{(I)}& 1D Conv Mixer (TAPIR)~\cite{doersch2023tapir} & 3  & 11.5 &  \underline{66.1} & \underline{78.0} & \underline{87.5} \\
\hlrow \textbf{(I\textit{})}& \ours-B (Ours) & 3 & 11.5 &  \textbf{67.8} & \textbf{79.6} & \textbf{89.9}\\

\bottomrule
  \end{tabular}
  }
  
  }
\vspace{-10pt}
\end{table}

\begin{figure}[t]
  \centering
\includegraphics[width=\linewidth]{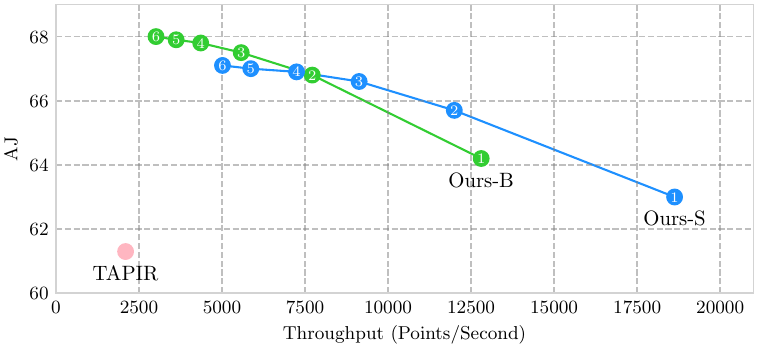}
\includegraphics[width=\linewidth]{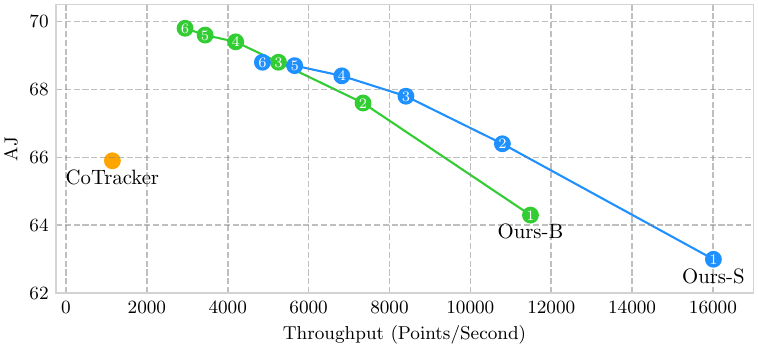}
\\
\vspace{-10pt}
\caption{\textbf{Results with a varying number of refinement iterations on TAP-Vid-DAVIS.} The number in the circle denotes the number of iterations. (up) In a 256\(\times\)256 resolution, compared to TAPIR~\cite{doersch2023tapir}, \ours achieves better performance in a \textit{single iteration} while being about \textit{9\(\times\) faster}. (below) In a 384\(\times\)512 resolution, compared to CoTracker~\cite{karaev2023cotracker}, \ours achieves comparable performance while being about \textit{9\(\times\) faster}.}
\label{fig:iterations}
\vspace{-10pt}
\end{figure}

\vspace{-10pt}
\subsubsection{Analysis on local correlation.}
In Table~\ref{tab:local-correlation}, we analyze the construction of our local correlation method, focusing on how we sample neighboring points around the query points rather than target points. \textbf{(I)} represents the performance of local 2D correlation, a common approach in the literature~\cite{harley2022particle,doersch2023tapir,karaev2023cotracker}. The performance gap between \textbf{(I)} and \textbf{(VI)} demonstrates the superiority of our 4D correlation approach over 2D.  \textbf{(II)} and \textbf{(III)} investigate the importance of calculating dense all-pair correlations within the local region. In \textbf{(II)}, we use randomly sampled positions for the query point's neighbors, while \textbf{(III)} uses a horizontal line-shaped neighborhood. Their inferior performance compared to \textbf{(IV)}, which samples the same number of points densely, emphasizes the value of our all-pair local 4D correlation. \textbf{(IV)} and \textbf{(V)} examine the effect of local region size. The gap between \textbf{(IV)} and \textbf{(V)}, supports our choice of region size. \textbf{(V)} represents our final model.

\vspace{-10pt}
\subsubsection{Ablation on position encoding of Transformer.}
In Table~\ref{tab:position-encoding}, we ablate the effect of relative position bias. With sinusoidal encoding~\cite{vaswani2017attention}, we observe significant performance degradation during inference (\textbf{I}) with variable length. In contrast, relative position bias demonstrates generalization to unseen sequence lengths at inference time (\textbf{II}). This approach eliminates the need for hand-designed chaining processes (\ie, sliding window inference~\cite{harley2022particle,karaev2023cotracker}) where window overlapping leads to computational inefficiency.

\vspace{-10pt}
\subsubsection{Ablation on the architecture of refinement model.} We verify the advantages of using a Transformer architecture over a Convolution-based architecture in Table~\ref{tab:architecture-ablation}. Our comparison includes the architecture proposed in Doersch et al.~\cite{doersch2023tapir}, which replaces the token mixing layer of MLP-Mixer~\cite{tolstikhin2021mlp} with depth-wise 1D convolution. We ensure a fair comparison by matching the number of parameters and layers between the models. Our Transformer-based model achieves superior performance. We believe this difference stems from their receptive fields: Transformers can achieve a global receptive field within a single layer, while convolutions require multiple stacked layers. Although convolutions can also achieve large receptive fields with lightweight designs~\cite{chen2017rethinking,dai2017deformable}, their exploration in long-range point tracking remains a promising area for future work.

\vspace{-10pt}
\subsubsection{Analysis on the number of iterations.}
We show the performance and throughput of our model, varying the number of iterations, in Fig.~\ref{fig:iterations}. We compare our model with TAPIR and CoTracker at their respective resolutions. Surprisingly, our model surpasses TAPIR even with a single iteration for both the small and base variants. With a single iteration, our small variant is about \(9\times\) faster than TAPIR. Compared to CoTracker, our model is about \(9\times\) faster at the same performance level.

\section{Conclusion}
We introduce \ours, an approach to the point tracking task, addressing the shortcomings of existing methods that rely solely on local 2D correlation. Our core innovation lies in a local all-pair correspondence formulation, combining the rich spatial context of 4D correlation with computational efficiency by limiting the search range.  Further, a  length-generalizable Transformer empowers the model to handle videos of varying lengths, eliminating the need for hand-designed processes. Our approach demonstrates superior performance and real-time inference while requiring significantly less computation compared to state-of-the-art methods. 

\section*{Acknowledgements} This research was supported by the MSIT, Korea (IITP-2024-2020-0-01819, RS-2023-00227592), Culture, Sports, and Tourism R\&D Program through the Korea Creative Content Agency grant funded by the Ministry of Culture, Sports and Tourism (Research on neural watermark technology for copyright protection of generative AI 3D content, RS-2024-00348469, RS-2024-00333068) and National Research Foundation of Korea (RS-2024-00346597).
\clearpage

\renewcommand{\thesection}{\Alph{section}}
\setcounter{section}{0}

\begin{center}
    \Large \bfseries Local All-Pair Correspondence for Point Tracking \\[5pt]
    \large --Supplementary Material--
\end{center}

\section{More Implementation Details}
For generating the Panning-MOVi-E dataset~\cite{doersch2023tapir}, we randomly add 10-20 static objects and 5-10 dynamic objects to each scene. The dataset comprises 10,000 videos, including a validation set of 250.
For the sinusoidal position encoding function~\cite{tancik2020fourier} \(\sigma(\cdot)\), we use a channel size of 20 along with the original unnormalized coordinate. This results in a total of 21 channels. For all qualitative comparisons, we use \ours-B model with a resolution of 384\(\times\)512.
\subsubsection{Details of the evaluation benchmark.}
We evaluate the precision of the predicted tracks using the TAP-Vid benchmark~\cite{doersch2022tap}. This benchmark comprises both real-world video datasets and synthetic video datasets.
\textbf{TAP-Vid-Kinetics} includes 1,189 real-world videos from the Kinetics~\cite{kay2017kinetics} dataset. As the videos are collected from YouTube, they often contain edits such as scene cuts, text, fade-ins or -outs, or captions. 
\textbf{TAP-Vid-DAVIS} comprises real-world videos from the DAVIS~\cite{pont20172017} dataset. This dataset includes 30 videos featuring various concepts of objects with deformations. 
\textbf{TAP-Vid-RGB-Stacking} consists of 50 synthetic videos~\cite{lee2021beyond}. These videos feature a robot arm stacking geometric shapes against a monotonic background, with the camera remaining static. In addition to the TAP-Vid benchmark, we also evaluate our model on the \textbf{RoboTAP} dataset~\cite{vecerik2023robotap}, which comprises 265 real-world videos of robot arm manipulation.

\begin{table}[h]
\centering
\vspace{10pt}
\caption{\textbf{Convolutional layer configurations for different model sizes.}}
\vspace{-10pt}
\label{tab:model_configs}
\begin{tabular}{l|ccc}
\toprule
Model & Channel Sizes & Kernel Size & Strides \\ \midrule
Small & (64, 128) & (5, 2) & (4, 2) \\
Base & (64, 128, 128) & (3, 3, 2) & (2, 2, 2) \\ 
\bottomrule
\end{tabular}
\end{table}
\subsubsection{Detailed architecture of local 4D correlation encoder.}
We stack blocks of convolutional layers, where each block consists of a 2D convolution, group normalization~\cite{wu2018group}, and ReLU activation. See Table~\ref{tab:model_configs} for details. For the small model, we use an intermediate channel size of (64, 128) for each block. For the base model, the intermediate channel sizes are (64, 128, 128) for each block. For every instance of group normalization, we set the group size to 16.

\subsubsection{Details of correlation visualization.}
For the correlation visualization in Fig. 3 of the main text, we train a linear layer to project the correlation embedding \(E_t^k\) into a local 2D correlation with a shape of 7\(\times\)7. This local 2D correlation then undergoes a softargmax operation to predict the error relative to the ground truth. We begin with the pre-trained model and train the linear layer for 20,000 iterations. For clarity, we bilinearly upsample the 7\(\times\)7 correlation to 256\(\times\)256.

\section{More Qualitative Comparison}
\begin{figure}[t]
  \centering
  \includegraphics[width=0.9\linewidth]{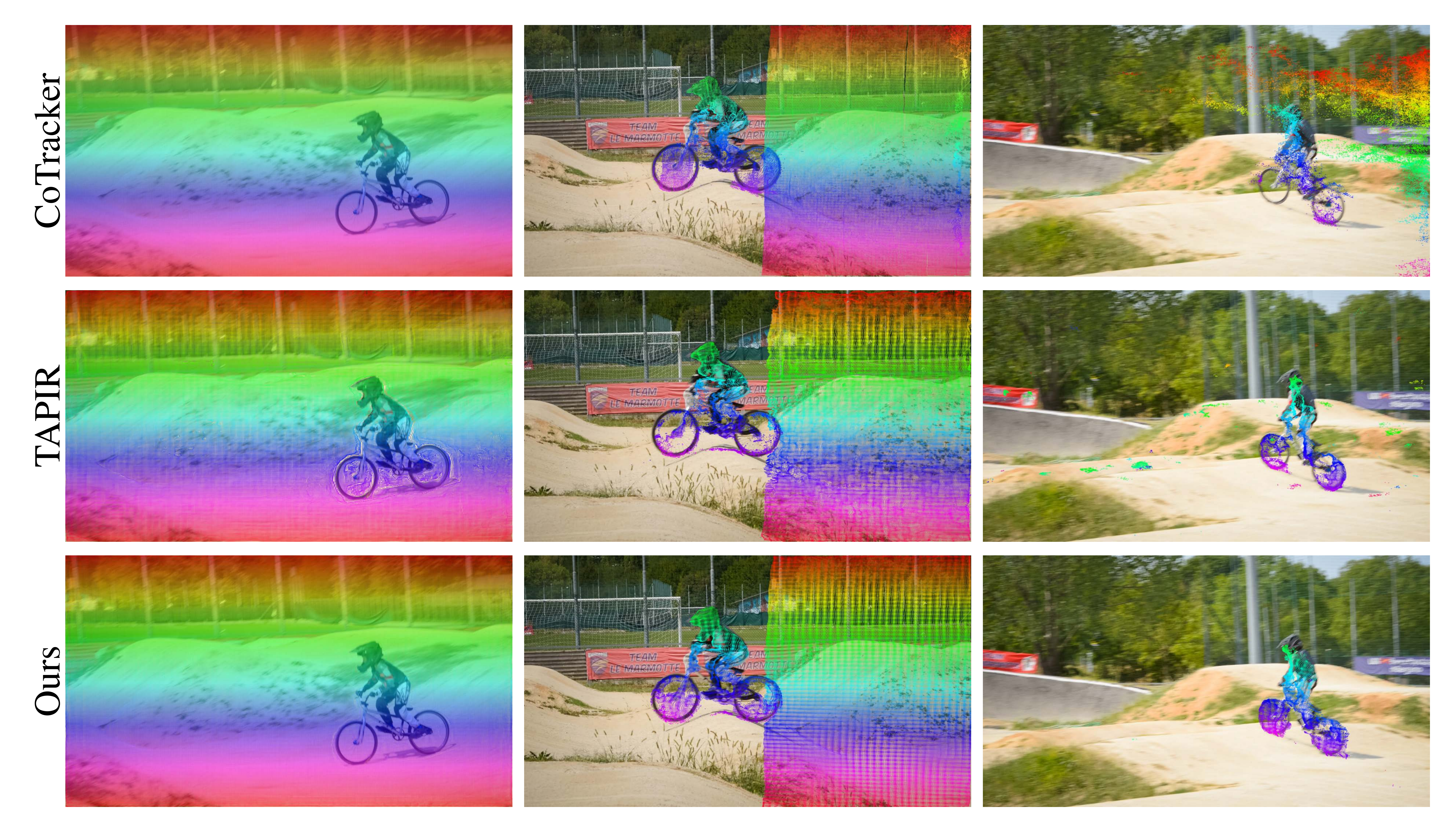}
  \includegraphics[width=0.9\linewidth]{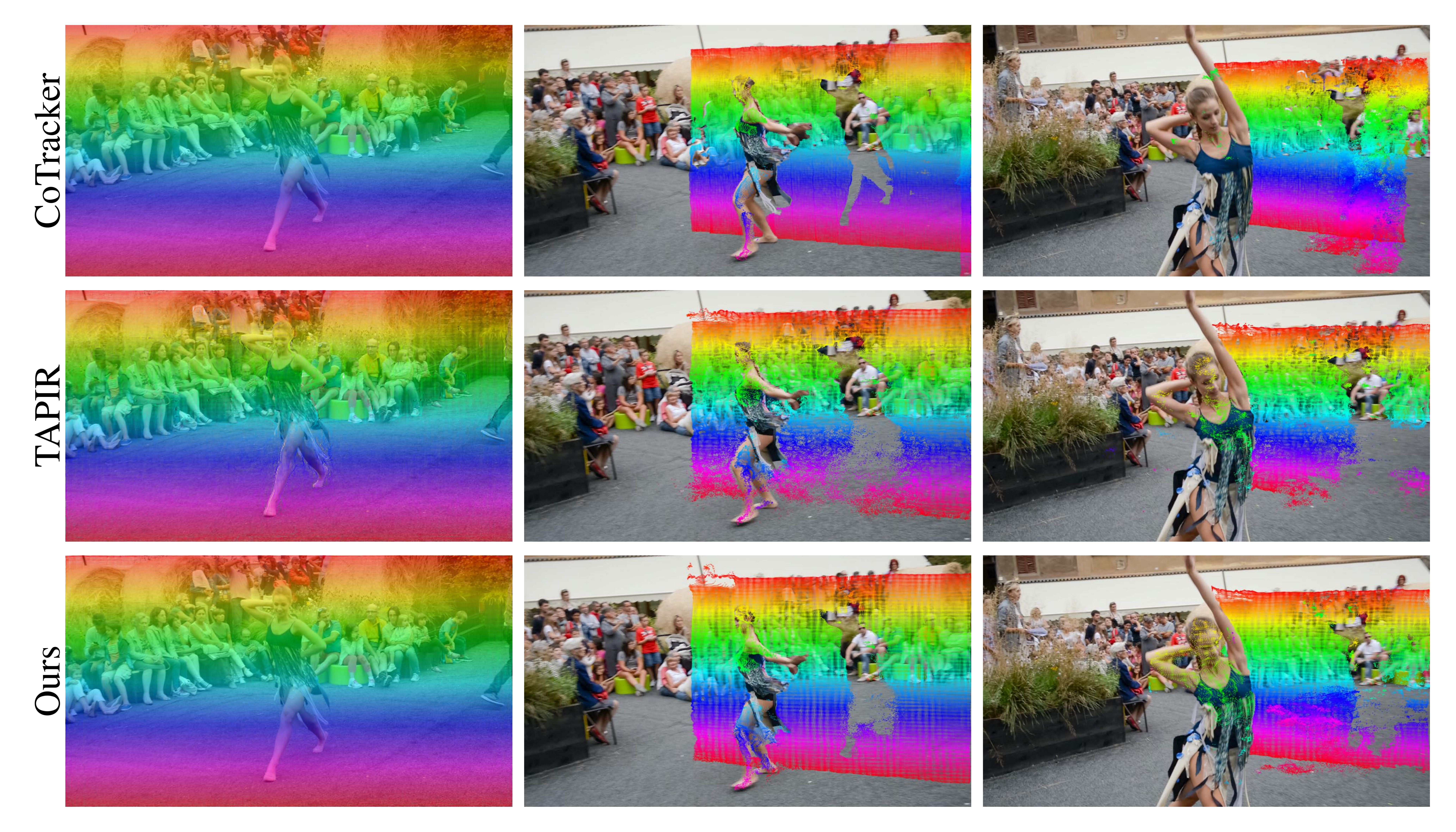}
  \includegraphics[width=0.9\linewidth]{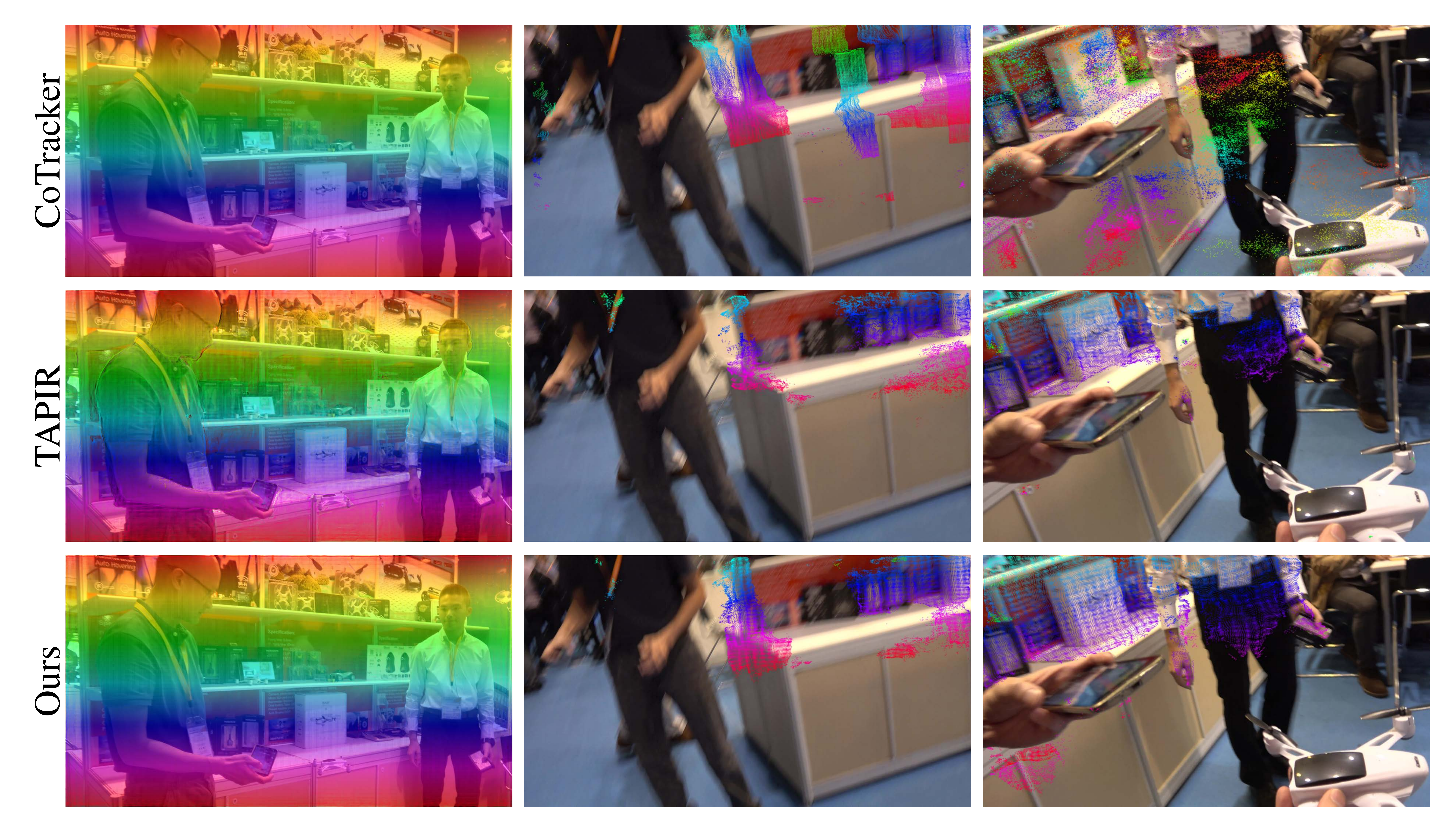}
  \vspace{-5pt}
  \caption{\textbf{Additional qualitative comparison with state-of-the-art~\cite{doersch2023tapir,karaev2023cotracker}.} }
  \label{fig:addtional_qual1}
\vspace{-10pt}
\end{figure}

\begin{figure}[t]
  \centering
  \includegraphics[width=0.9\linewidth]{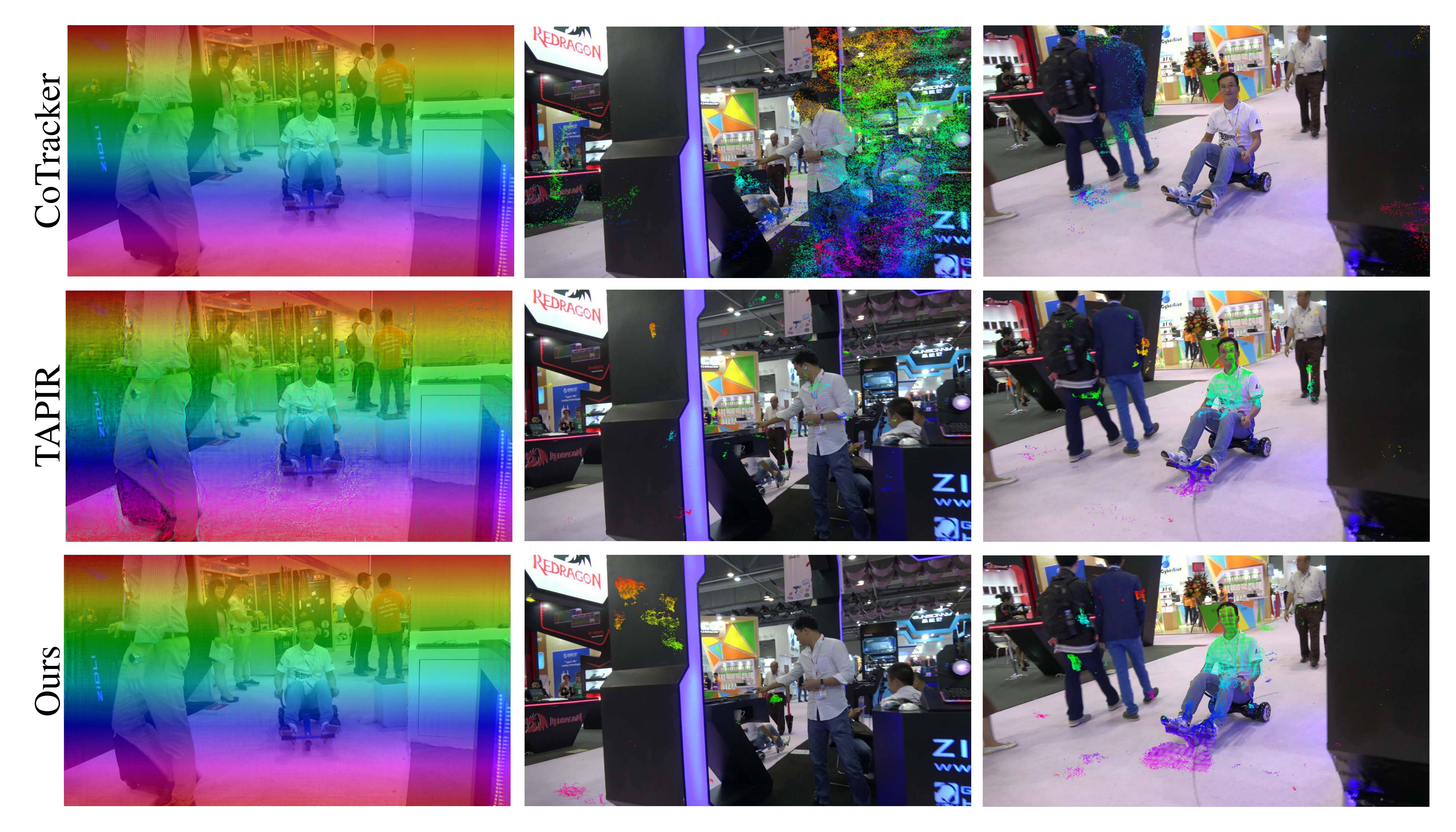}
  \includegraphics[width=0.9\linewidth]{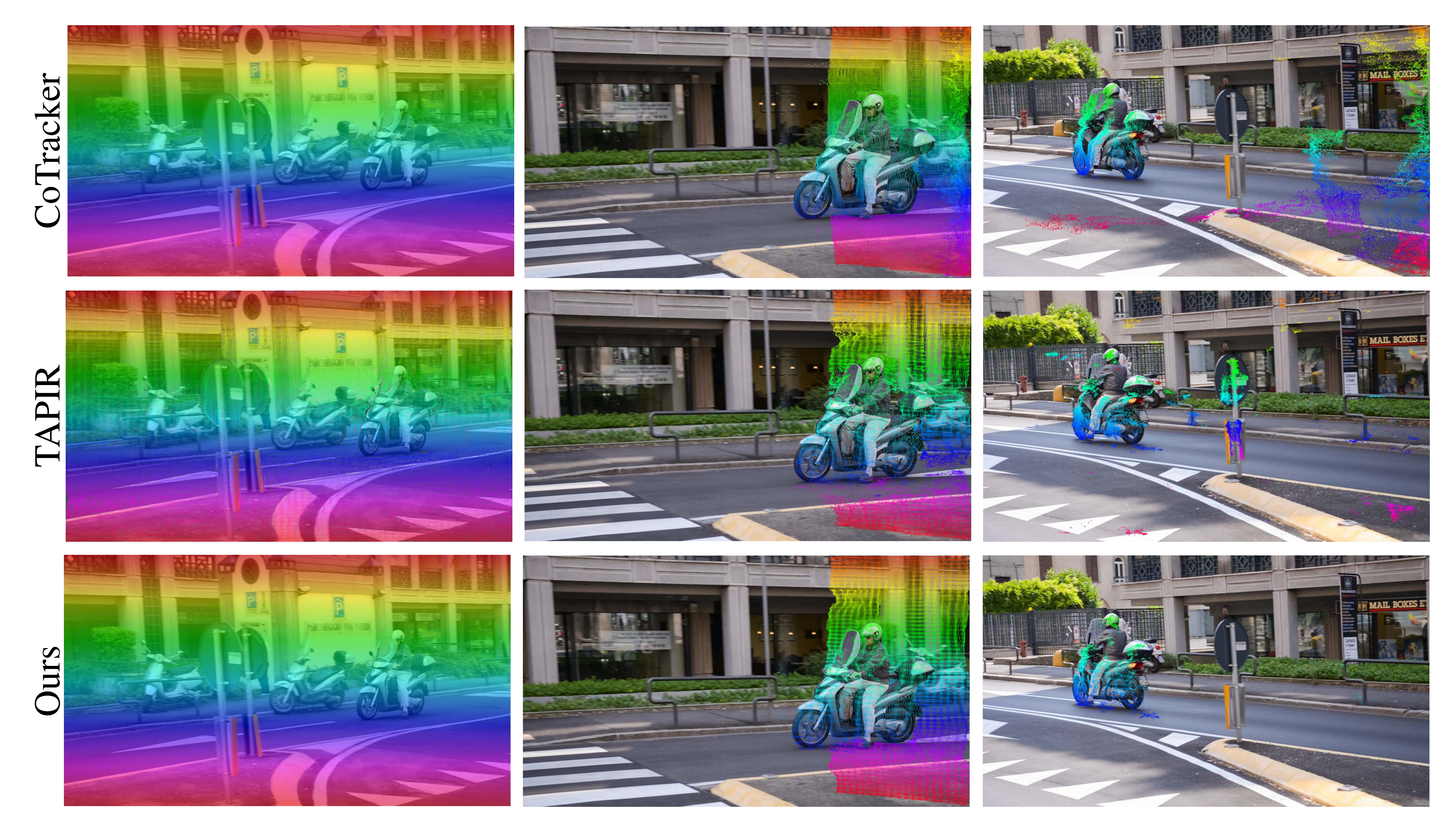}
  \includegraphics[width=0.9\linewidth]{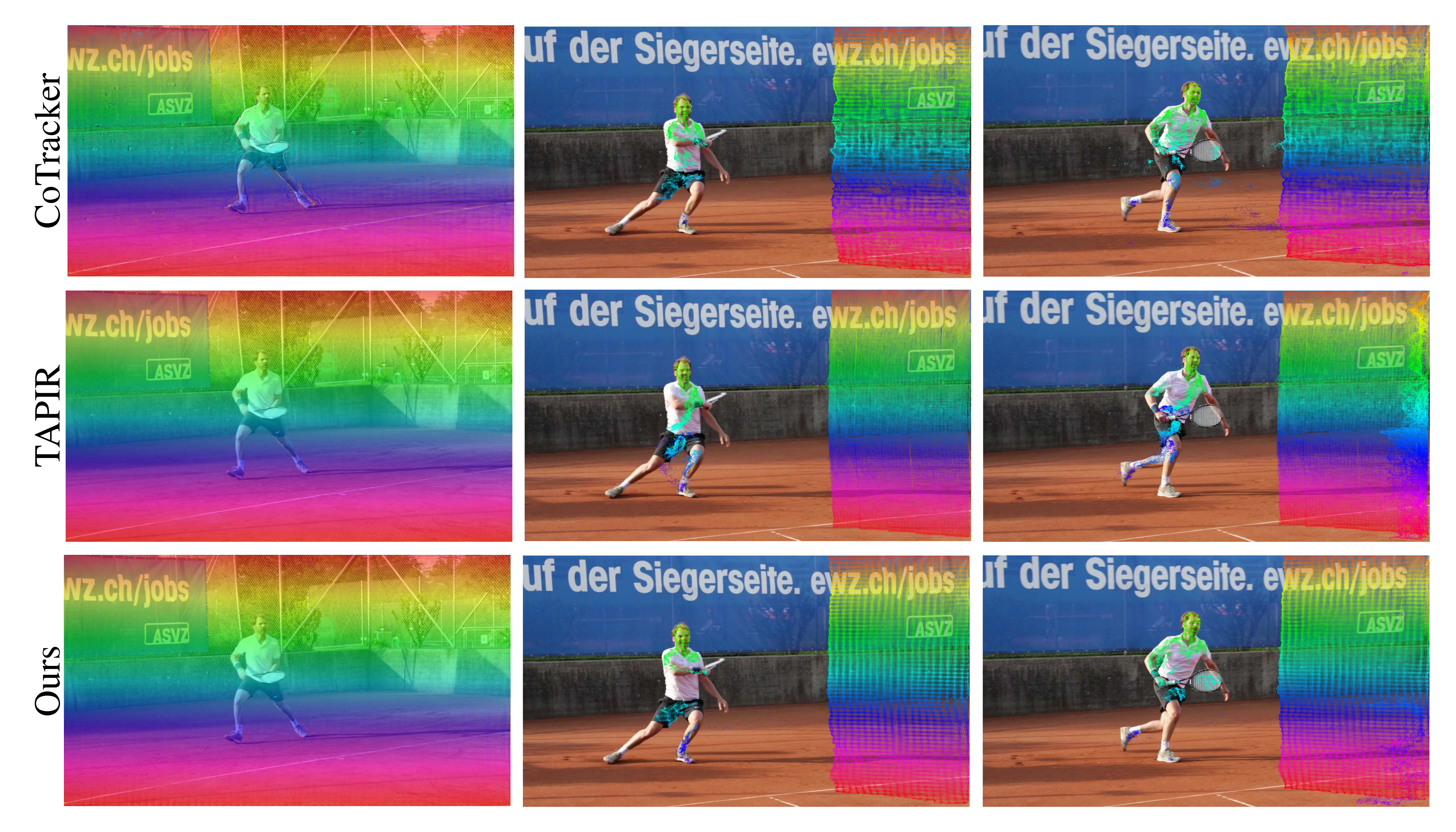}
  \vspace{-5pt}
  \caption{\textbf{Additional qualitative comparison with state-of-the-art~\cite{doersch2023tapir,karaev2023cotracker}.} }
  \label{fig:addtional_qual2}
\vspace{-10pt}
\end{figure}

We provide more qualitative comparisons to recent state-of-the-art methods~\cite{doersch2023tapir,karaev2023cotracker} in Fig.~\ref{fig:addtional_qual1} and Fig.~\ref{fig:addtional_qual2}. Our model establishes accurate correspondences in homogeneous areas and on deforming objects, demonstrating robust occlusion handling even under severe occlusion conditions.

\clearpage

%
%
\bibliographystyle{splncs04}
\bibliography{main}
\end{document}